\documentclass[letterpaper, 10 pt, conference]{ieeeconf}  

\IEEEoverridecommandlockouts                              

\usepackage{graphicx}
\usepackage{subcaption}
\usepackage{rotating}
\usepackage{amsmath}
\usepackage{float}
\usepackage[usenames,dvipsnames]{xcolor}
\usepackage{tikz}
\usepackage{amsmath, amssymb, amsfonts}
\usepackage[ruled,vlined,linesnumbered]{algorithm2e}
\usepackage{array}
\usepackage{algorithmic}
\usepackage{graphbox}
\usepackage{wrapfig}
\usepackage{makecell}
\usepackage{booktabs}
\usepackage{url}
\usepackage{balance}
\usepackage{cite}
\usepackage{multirow}
\usepackage{xcolor}
\usepackage{colortbl}
\usepackage{hyperref}

\newcommand{\methodname}{DDP\xspace} 

\title{\LARGE \bf Dino-Diffusion Modular Designs Bridge the Cross-Domain Gap in Autonomous Parking}

\author{Zixuan Wu${^{1}}$, Hengyuan Zhang${^2}$, Ting-Hsuan Chen${^3}$, Yuliang Guo$^{2\dagger}$, David Paz$^{2}$, \\Xinyu Huang$^{2}$, and Liu Ren$^{2}$
\thanks{$^{1}$ Zixuan Wu is with Bosch Research North America, Bosch Center for AI (BCAI) when the work is done in his internship and Institute for Robotics and Intelligent Machines (IRIM), Georgia Institute of Technology, Atlanta, GA 30332, USA {(Corresponding Author) \tt\small zwu380@gatech.edu}.} %
\thanks{$^{2}$ Hengyuan Zhang, Yuliang Guo, David Paz, Xinyu Huang and Liu Ren are with Bosch Research North America and Bosch Center for AI (BCAI), 384 Santa Trinita Ave, Sunnyvale, CA 94085, USA  {\tt\small \{henry.zhang2, yuliang.guo2, david.pazruiz, xinyu.huang, liu.ren\}@us.bosch.com}.}%
\thanks{$^{3}$ Ting-Hsuan Chen is with the Department of Computer Science, University of Southern California, 3551 Trousdale Parkway, Los Angeles, CA 90089, USA {\tt\small tchen783@usc.edu}.}%
\thanks{$^{\dagger}$ Tech lead.}
}

\begin{document}

\maketitle
\thispagestyle{empty}
\pagestyle{empty}

\begin{abstract}    
Parking is a critical pillar of driving safety. While recent end-to-end (E2E) approaches have achieved promising in-domain results, robustness under domain shifts (e.g., weather and lighting changes) remains a key challenge. 
Rather than relying on additional data, in this paper, we propose Dino-Diffusion Parking (\methodname), a domain-agnostic autonomous parking pipeline that integrates visual foundation models with diffusion-based planning to enable generalized perception and robust motion planning under distribution shifts. 
We train our pipeline in CARLA at regular setting and transfer it to more adversarial settings in a zero-shot fashion. Our model consistently achieves a parking success rate above 90\% across all tested out-of-distribution (OOD) scenarios, with ablation studies confirming that both the network architecture and algorithmic design significantly enhance cross-domain performance over existing baselines. Furthermore, testing in a 3D Gaussian splatting (3DGS) environment reconstructed from a real-world parking lot demonstrates promising sim-to-real transfer\footnote[4]{\ \ Code is at \url{https://github.com/ChampagneAndfragrance/Dino_Diffusion_Parking_Official}}.
\end{abstract}
\section{Introduction}

Autonomous driving has advanced rapidly in recent decades, with features now present in nearly 60\% of new vehicles worldwide~\cite{statista2025autonomous}. These systems perceive the environment through multiple sensors, make decisions from vehicle states, and execute control with reduced human input~\cite{nhtsa2025automated}. Unlike open-road driving, autonomous parking poses distinct challenges: precise maneuvering in constrained spaces, frequent directional changes, and complex low-speed navigation~\cite{zhang2024reward}. Notably, parking contributes to 20\% of all U.S. vehicle accidents, with 91\% linked to backing maneuvers, emphasizing the critical role of accurate perception, planning, and control~\cite{aaa2016parking}.

Some classic research on the autonomous parking assume an accurate map information and focus more on the trajectory generation with search-based methods \cite{zhao2024automatic, leu2022improved} and execution with optimal or model predictive controls \cite{leu2022autonomous, shen2021collision}. To make it a fully practical parking pipeline, people also research how to build Bird’s-Eye View (BEV) representations from the perspective surround-view images \cite{yang2021towards, philion2020lift, li2024bevformer} and utilize them for perception tasks. 

Recent trends are shifting to end-to-end (E2E) models \cite{rathour2018vision, hu2022st, E2EAPA}. Compared with the rule-based map segmentation and motion planning \cite{yang2021towards}, the E2E methods unify sensor data and motion planning to leverage diverse real world data for generalization and avoid tedious handcrafted computational processes. A concurrent work CAA-Policy~\cite{Chen2025caapolicy} further improves the E2E parking frameworks~\cite{E2EAPA} by back-propagating an attention map from control to perception, which leads to a much more robust E2E parking policy under the in-domain test. 

\begin{figure}
    \centering
    \includegraphics[width=1.0\linewidth]{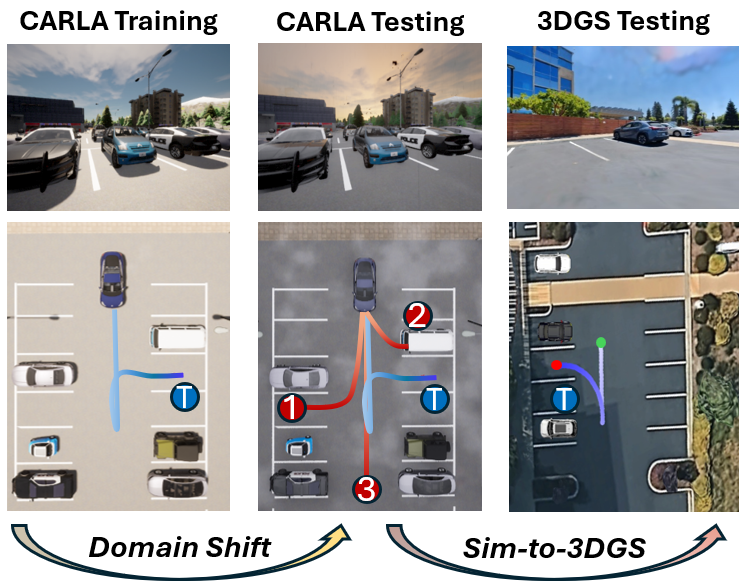}
    \caption{\textbf{Zero-Shot Domain Generalization:} Our method addresses the challenge of training a parking policy under default weather conditions (e.g., CARLA sunny-noon, left) and deploying it zero-shot in other weather or illumination settings (middle) or in a 3DGS environment reconstructed from the real world (right). Our method (blue trajectories indicated by 'T') overcomes the cross-domain failures (red trajectories) observed in prior in-domain E2E pipelines and demonstrates promising zero-shot transfer to 3DGS testing.}
    \label{fig:idea}
    \vspace{-5mm}
\end{figure}

Although E2E methods demonstrate promising performance improvements, they often require a close match of the training and testing distribution. It has been shown that the domain gaps in visual appearance (textures, lighting, weather) will degrade the policy transferability  \cite{mueller2018driving} and the compounding of small deviations may put the vehicle in unfamiliar states, leading to cascading mistakes in behavior cloning \cite{codevilla2019exploring}. Solutions to these problems include the domain randomization \cite{domain_randomization}, modularity with semantic abstraction \cite{mueller2018driving}, and embedding tokenization and regularization \cite{yasarla2025roca}. In the recent years, the rising of the visual foundation models (VFM) provide us an alternative way other than the expensive road data collection or tedious loss design. The VFM architectures \cite{dosovitskiy2020image, caron2021dino, oquab2023dinov2} pretrained on massive video datasets demonstrate strong transferability, robustness, and zero-shot capabilities compared with task-specific models in downstreaming tasks \cite{kirillov2023segment, depthanything}. Surprisingly, there is rarely work discussing how the VFMs can help on the cross-domain transfer of the parking policy. 

In this paper, we address an important yet under-explored question: how to manage significant distribution shifts between the expert training domain and the deployment domain without relying on additional data collection. Specifically, we identify three major factors that limit cross-domain performance: visual disturbance (e.g. changes in illumination, weather, or object appearance), domain overfitting (i.e. perception modules overfits to domain-specific rather than generalizable cues) and compounding error (i.e. accumulation of the error in the action space). Existing E2E parking approaches \cite{E2EAPA, li2024parkinge2e, Chen2025caapolicy} struggle to adapt under such shifts, often producing out-of-distribution (OOD) trajectories that result in deviations, collisions, or outright parking failures (see red trajectories in Figure \ref{fig:idea}). 

To address these limitations, we design a decoupled and cascaded pipeline that leverages a VFM and a diffusion approach for trajectory modeling. The approach also leverages data augmentation techniques and achieves significant generalization improvements for cross-domain tasks. Our contribution can be listed as follows:

\begin{enumerate}
    \item To our best knowledge, this is the first work to investigate the cross-domain autonomous parking and achieves zero-shot transferability.
    \item We pioneer a modular autonomous parking pipeline that decouples generalizable perception (with VFM and data relabeling), robust planning (with diffusion) and accurate tracking (with Stanley control) such that the representations will not overfit to the training set. 
    \item Our framework consistently and significantly outperforms the state-of-the-art in the CARLA cross-domain benchmarking. We also show the feasibility of our pipeline in a 3D Gaussian Splatting (3DGS) environment reconstructed from a real-world parking lot. 
\end{enumerate}

\begin{figure*}
    \centering
    \includegraphics[width=0.9\linewidth]{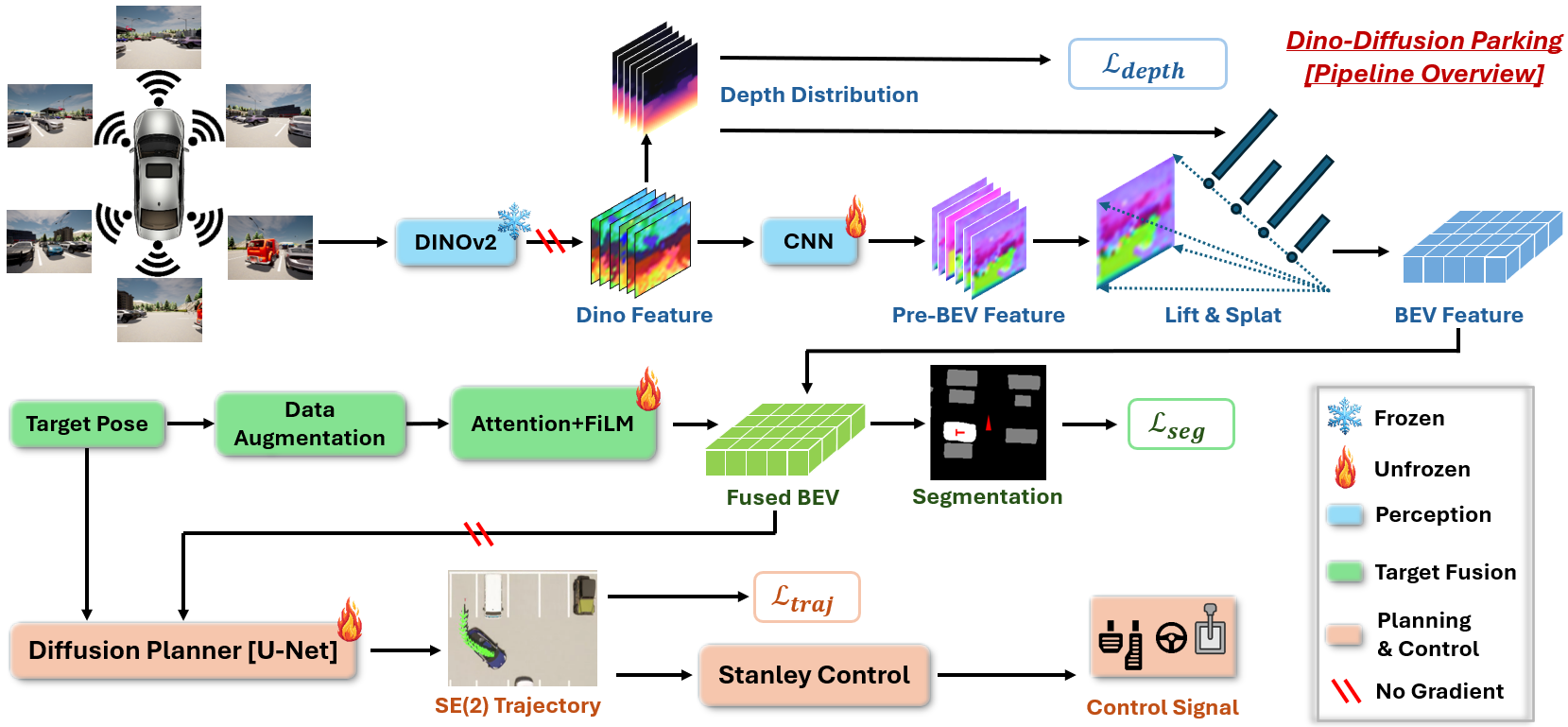}
    \caption{\textbf{Dino-Diffusion Parking (\methodname) Pipeline:} We use DINOv2 to generate robust visual features and transform features into vehicle BEV. Then we relabel the target parking spot (in training only) and fuse with BEV. Finally, the diffusion planner employs a U-Net conditioned on the target pose and the fused feature to predict the future trajectory in $\mathrm{SE(2)}$ and execute it with Stanley control. The supervision signal is from ground truth depth, segmentation and expert trajectories, and it should be noted that \textit{the perception and planning are decoupled where the gradient doesn't pass from diffusion to fused BEV.}}
    \label{fig:full_pipeline}
    \vspace{-5mm}
\end{figure*}
\section{Related Work}
\label{sec:related_works}
\subsection{Autonomous Parking}
Autonomous parking system aims to have a vehicle precisely park to a target slot in a cluttered parking lot at a low-speed\cite{olmos2025overview}. Therefore, it requires a robust surrounding perception, a long-horizon motion planning and an accurate command execution. The perception module usually abstracts the surrounding scenarios into Bird’s-Eye View (BEV), which is a top-down occupancy map in ego-vehicle coordinates~\cite{philion2020lift, li2024bevformer, liang2022bevfusion}. 

E2E motion planning~\cite{bojarski2016end, E2EAPA, li2024parkinge2e, hu2022st, rathour2018vision} learns to map raw sensor inputs directly to trajectories or control commands (i.e. brake, throttle, steer, gear). Early works used MLPs~\cite{pomerleau1988alvinn} or CNNs~\cite{bojarski2016end, rathour2018vision} without intermediate representations, while BEV-conditioned approaches~\cite{E2EAPA, hu2022st, rathour2018vision} partition perception and planning, improving modularity and generalization. A recent work~\cite{yasarla2025roca} addresses cross-domain transfer via latent trajectory tokens with Gaussian Process regularization. In contrast, we tackle transfer from the pipeline design side, leveraging visual foundation models, data relabeling, and diffusion trajectory modeling for stronger generalization.

\subsection{Domain Transfer and Foundation Models}

Domain transfer from changes in weather, lighting, and layouts remains a major challenge in autonomous driving. While sensor fusion improves robustness~\cite{liu2022bevfusion, lin2024rcbevdet}, its cost motivates low-cost, vision-based solutions. Data Augmentation improves robustness in general~\cite{li2023domain}, but often requires domain knowledge or data. Training with diverse weather data has been proven to be the most effective approach~\cite{Gupta2024robustweather}, however, the data volume will unaffordably scale up especially when there is downsteaming tasks like parking.

The rise of vision foundation models (VFMs) marks a paradigm shift toward universal visual understanding~\cite{caron2021dino, oquab2023dinov2, kirillov2023segment, depthanything}. They offer strong generalization by learning from large-scale, diverse data. For instance, DINOv2~\cite{oquab2023dinov2}, trained on 142M curated images, provides versatile features across distributions, while SAM~\cite{kirillov2023segment}, trained with over 1B masks, offers strong generalization with flexible prompting. We investigate leveraging DINOv2’s semantic and geometric understanding to enable consistent perception across varying conditions in the parking without any extra data collection.

\subsection{Imitation Based Motion Planning and Control} 

Interestingly, one of the earliest and most recognized works in imitation learning \cite{pomerleau1988alvinn} is also in autonomous driving, training a 3-layer fully connected network to map raw images to steering commands. However, later work \cite{ross2010efficient} showed that behavior cloning suffers from compounding errors in sequential decision-making. To address this drifting problem, Dataset Aggregation (DAGGER) \cite{ross2011reduction} was proposed, allowing the policy to iteratively query the expert during training and achieving strong performance in the Super Tux Kart Star Track benchmark.

Generative models have since advanced imitation learning by enabling richer scene understanding and sequential modeling. GAIL~\cite{ho2016generative, fu2017learning} frames policy learning as adversarial training to match expert occupancy distributions, though at the cost of instability~\cite{li2023learning}. More recently, transformers and diffusion models explicitly predict future action sequences, mitigating drift and improving stability. Decision Transformer~\cite{chen2021decision}, ACT~\cite{zhao2023learning}, and diffusion policies~\cite{janner2022planning, chi2023diffusionpolicy} demonstrate strong results in manipulation tasks. Diffusion has also been applied to navigation~\cite{jiang2023motiondiffuser, zheng2025diffusion, wu2025learningwheelchairtennisnavigation}, but these tasks rely on structured inputs rather than raw images and do not address the tight spatial constraints of autonomous parking as we do.
\section{Methodology}
In this section, we will describe how to build our autonomous parking pipeline, Dino-Diffussion Parking (\methodname), which is robust to the distribution shift across domains. In \S\ref{sec:method_robust_perception}, we employ VFMs to achieve a generalized surrounding perception. The following \S\ref{sec:augmentation} further illustrates data augmentation through simply relabeling the target can help the the robust parking goal recognition. In \S\ref{sec:motion_planning}, we design a diffusion based motion planning within Cartesian space synergy with a pose control, which has been shown to suffer less from compounding error \cite{chi2023diffusionpolicy}. The full pipeline overview is visualized in Figure \ref{fig:full_pipeline}.

\subsection{Robust Surrounding Perception through VFM}
\label{sec:method_robust_perception}
To obtain a robust representation of the car surroundings, we need to guarantee the similar camera image embeddings can be derived from the visually equivalent parking scenarios. Here \textit{visually equivalent} in our parking task means the visual disturbance should not change how the vehicle behaves. For example, the vehicle should plan similar paths from the road to the target parking slot if only the weather condition or the types of other parked cars are changed. 

The consistency in visual embeddings can lead to a stable conditioning for the following information fusion and motion modules in cross-domain tasks. However, it is difficult to perform zero-shot transferring to the unseen scenarios, especially for the E2E vision-action pipeline \cite{E2EAPA} where the whole network overfits to the training set. Therefore, we consider to use an advanced VFM DINOv2 as the image processing backbone, which has been pre-trained with the diverse world data. During our training, we freeze the DINOv2, lift the patched features into 3D spaces, and splat it onto the BEV \cite{philion2020lift}. We show our method makes the scene understanding more generalizable (see \S\ref{sec:res_perception}). Our algorithm can be summarized as Algorithm \ref{alg:pseudocode_robust_bev}.

\SetCommentSty{mybluecomment}
\newcommand{\mybluecomment}[1]{\textcolor{gray}{\small\textit{#1}}}
\SetKwComment{Comment}{$\lhd$ }{}
\begin{algorithm}[h]
\SetAlgoLined
\KwData{$I$ (camera images), $\phi_{dino}$ (DINOv2 model), $F$ (patched DINO features), $\phi_f$ (convolutional layers for features), $\phi_d$ (convolutional layers for depth), $\phi_b$ (BEV generator), $n_I$ (the number of surrounding cameras), $n_F$ (the number of patches)}
\Comment{Generate DINOv2 patched features} 
$F=\phi_{dino}(I)=\{F_{ij}\}_{i,j=1,1}^{n_I,n_F}=\{F_j\}_{j=1}^{n_I\times n_F}$\; \label{eq:dino_feature}
\Comment{Feature post-process and depth generation} 
$F=\phi_f(F), F_d=\phi_d(F)$\; \label{eq:post_process}
\Comment{Generate Robust BEV \cite{philion2020lift}}
$F_b=\phi_b(F, F_d, CamParameter)$\;  \label{eq:bev_generation}
\caption{Robust BEV Generation}
\label{alg:pseudocode_robust_bev}
\end{algorithm}
 
\subsection{Generalized Target Fusion via Relabeling}
\label{sec:augmentation}
We already derived a robust BEV in \S\ref{sec:method_robust_perception}, however, the full perception module also needs to fuse the target slot information in the feature embeddings. The target slot is specified by the user with target pose $P_{tgt}$ and invisible in the camera scenes, and it is equivalently important to have the target pose robustly merged into the BEV feature. 

In our cross-domain test, we find the perception module trained in the original setting can recognize many slots near the vehicle position as the goal regardless of the input target pose (see \S\ref{sec:data_augmentation}). It shows the target representation can unexpectedly overfit to some domain specific information and the expert success parking pattern rather than focus on the user input. This problem has been ignored by the previous works \cite{E2EAPA, Chen2025caapolicy} and will cause troubles when the sensor information is out-of-distribution.

A potential solution is to collect more non-target-parking data as an augmentation to the expert-only success trajectories, however, it will introduce considerable overhead in data-collection and training. Therefore, we design a hindsight target parking relabeling method for data augmentation: We manually perturb the target pose $P_{tgt}$, relabel the ground truth segmentation map $I_{s}$, and pair them together. During the training, we iteratively train with the expert pairs and relabeled pairs such that the training is sufficient and generalizable to all distribution. In addition, we use a Feature-wise Linear Modulation (FiLM) \cite{perez2018film} structure to modulate the BEV feature map with target pose conditioning before passing to the downstreaming blocks. It is more feasible than learning an attention map \cite{Chen2025caapolicy} from the back-propagation gradient to filter the BEV, since the gradient is cut between the perception and planning  in our modular training (see Figure \ref{fig:full_pipeline}). This part of algorithm can be summarized as Algorithm \ref{alg:pseudocode_robust_target}.

\begin{algorithm}[h]
\SetAlgoLined
\KwData{$F_b$ (BEV features), $P_{tgt}$ (target pose), $I_{s}$ (GT segmentation map), $N$ (training epochs), $T_p$ (training period), $T_{exp}$ (expert epochs in a period), $\phi_i$ (relabeling function), $\phi_s$ (segmentation head), $F_s$ (segmentation map)}
\For{$k \gets 1$ \textbf{to} $N$}
{
    \Comment{Periodically train with relabeling the target} 
    \If{$k\ \mathbf{mod}\ T_p < T_{exp}$} 
    {
        $\hat{P}_{tgt} = P_{tgt} + \mathcal{N}(\mu_a, \sigma_a)$\; 
        $\hat{I}_s = \phi_i(I_{s}, \hat{P}_{tgt})$
    }
    \Else
    {
        $\hat{P}_{tgt} = P_{tgt}$\;
        $\hat{I}_s = I_{s}$ 
    } 
    \Comment{Fuse target info with cross attention and FiLM} 
    $F_b = \mathbf{CrossAttn}(F_b, \hat{P}_{tgt})$\;
    $\gamma, \beta = \mathbf{MLP}(F_b)$, $F_b = \gamma \cdot F_b + \beta$\;
    \Comment{Generate the segmentation map from fused BEV} 
    $F_s = \phi_s(F_b)$

}
\label{eq:bev_generation2}
\caption{Generalized Target Tracking}
\label{alg:pseudocode_robust_target}
\end{algorithm}

\subsection{Cartesian Space Motion Planning with Diffusion Model}
\label{sec:motion_planning}
The previous E2E parking pipeline work conducts the motion planning directly in the terminal control space (i.e. throttle, brake, steering, and gear) \cite{E2EAPA}. However, recent work discovers that the motion plannning in Cartesian space in conjunction with pose control can help to decrease the compounding error \cite{chi2023diffusionpolicy}. When the authors in \cite{E2EAPA} test the sim-to-real transfer of their policy \cite{li2024parkinge2e}, they also follow the planning-control decoupled pattern.

We build the motion planning as a diffusion sequential modeling problem in $\mathrm{SE(2)}$ where the state is described by the ego-centric coordinates and heading angle $x, y$ and $\theta$, which is different from previous navigation works \cite{jiang2023motiondiffuser, wu2025learningwheelchairtennisnavigation} only diffusing waypoint location information $x, y$. Such heading information is important since the parking tasks require a precise alignment of the vechicle to the parking slot through directional waypoints. We use the diffusion model as a local motion planner conditioned on the most updated fused BEV embeddings, segmentation and target pose. Our algorithm can be summarized as Algorithm \ref{alg:pseudocode_diffusion_motion}.

\begin{algorithm}[t] 
\SetAlgoLined
\KwData{$\mathcal{T}_v = \{x_i, y_i, \theta_i\}_{i=1}^{T}$ (vechicle trajectory), $T_d$ (waypoint downsample rate), $N_w$ (waypoint number on each plan)} 
\Comment{Segment trajectory into pieces with $T_d\cdot N_w$ waypoints}
$\mathcal{T}_{v}^{sub} = \mathbf{Segment}(\mathcal{T}_v, T_d\cdot N_w) \label{eq:cut_traj}$\;
\Comment{Downsample trajectories every $T_d$ timesteps}
$\mathcal{T}_{v}^{sub} = \mathbf{Downsample}(\mathcal{T}_{v}^{sub}, T_d) \label{eq:downsample_traj}$\;
\While{Training iteration}
{ 
\Comment{Training trajectory $\tau$, diffusion step $i$, noise $\epsilon$}
$\tau_0, m \sim \mathcal{T}^{sub}; i \sim \mathcal{U}(1, T), \epsilon \sim \mathcal{N}(0, I);$
\label{eq:draw_traj} \\
\Comment{Diffusion forward pass: apply scheduled noise}
$\tau_i \leftarrow \sqrt{\bar{\alpha}_i} \tau_0 + \sqrt{1 - \bar{\alpha}}_i \epsilon$\;\label{eq:schedule_noise}
\Comment{Update denoising model $s_{\theta}$ conditioned on $F_b, F_s$}
$\mathcal{L}(\theta) = \| \epsilon - s_\theta(\tau_i, i, F_b, F_s))\|_2^2$\; \label{eq:gradient_descend}
}
\Comment{Start the inference stage}
$\tau^T \sim \mathbf{N}(0, I)$\Comment*[r]{Sample from Gaussian noise} \label{eq:noise}
\For{all $i$ from $T$ to $1$}{ \label{line:for}
\Comment{Denoise the trajectory for one step}
$(\mu^i, \Sigma^i) \leftarrow s_{\theta}(\cdot), \Sigma_{\theta}(\tau^{i}) \tau^{i-1} \sim \mathcal{N}(\mu^i, \Sigma^i)$\; \label{eq:denoise}
\Comment{Sample the denoised trajectory}
$\tau^{i-1} \sim \mathcal{N}(\mu^i, \Sigma^i);$\label{eq:sample}
 }
\caption{Diffusion Local Planner}
\vspace{-1mm}
\label{alg:pseudocode_diffusion_motion}
\end{algorithm}

\begin{figure}
    \centering
    \includegraphics[width=0.65\linewidth]{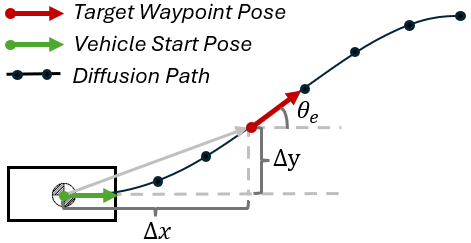}
    \caption{\textbf{Tracking with Stanley Control:} Given the trajectory from the diffusion planner, the controller minimizes the heading error and distance error with a reference point.} 
    \label{fig:stanley}
    \vspace{-5mm}
\end{figure}

\begin{figure*}
    \centering
    \includegraphics[width=0.8\linewidth]{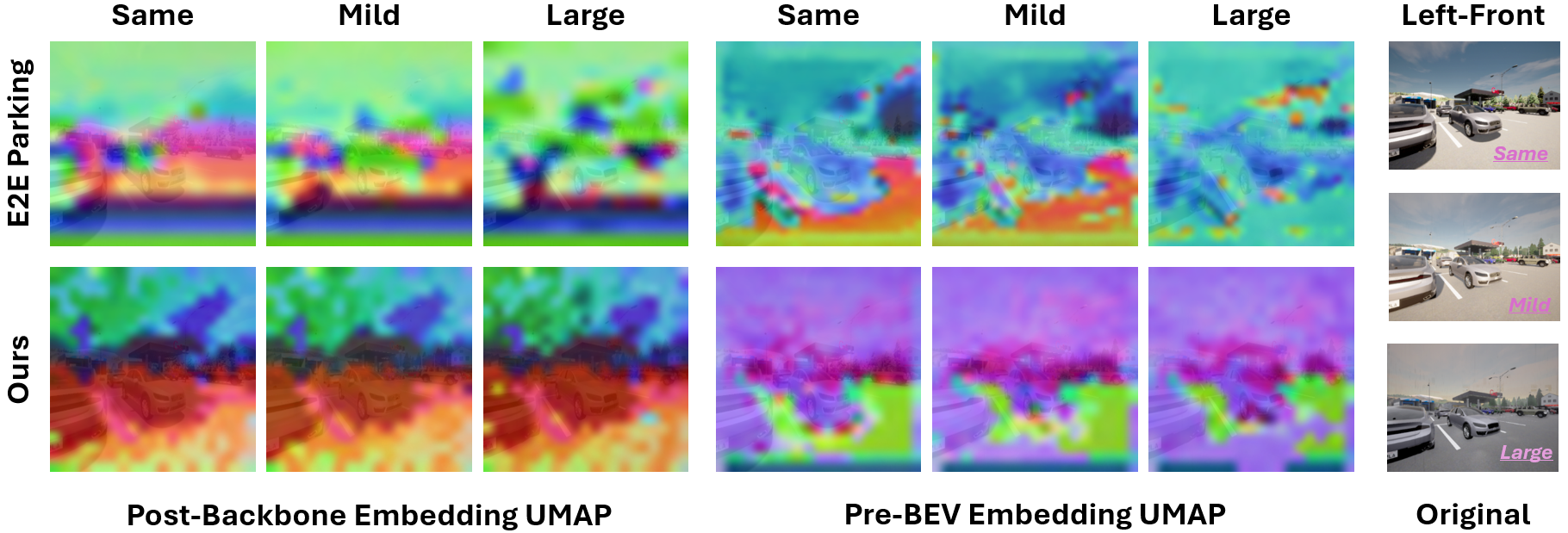}
    \caption{\textbf{Robust Features from Vision Foundation Models:} We utilize UMap~\cite{2018arXivUMAP} to visualize the features at two stages, directly from the backbone and before they are projected into BEV. Compared to the features from the baseline model, our features from DINOv2~\cite{oquab2023dinov2} backbone are more consistent across different domains, highlighting the potential to reduce domain gap with foundation models in perception.}
    \label{fig:robust_perception}
    \vspace{-5mm}
\end{figure*}

\subsection{Accurate Trajectory Tracking from Stanley Control}
The Stanley control, dated back to \cite{thrun2006stanley}, has been commonly used in the autonomous vehicle trajectory tracking. It is a steering control method containing two terms: 1) heading error to align the vehicle and the trajectory directions, and 2) cross-tracking error to drag the vehicle to the closest point on the trajectory. 

Since our motion plan is ego-centric trajectory (see Figure \ref{fig:stanley}), we can simplify the Stanley control and make the target point and error weights more flexible. Let's say the vehicle selects the $n$-th waypoint ahead as the target reference. Assume $\theta_e$ is the heading error between the vehicle and the target waypoint, $e_y$ is the lateral cross-track error, $v$ is the vehicle’s forward velocity, $k$ is a positive control gain, the controller steering angle $\delta$ can be computed as:
\begin{equation}
\delta = \theta_e + \arctan\!\left( \frac{k \cdot e_y}{v} \right) \approx k_e \cdot \theta_e + \frac{k_y}{v} \cdot \arctan(\frac{\Delta y}{\Delta x})
\label{eq:stanley}
\end{equation}
Our result in \S\ref{sec:result} shows it is extremely suitable to use diffusion motion planning in conjunction with Stanley control in the autonomous parking tasks, since such combination leverages the strengths of both data-driven planning and model-based control: the diffusion model outputs surrounding-aware consistent waypoint sequences, while the controller ensures accurate low-level tracking in the tight parking lot.

\begin{table}[t]
\centering
\caption{\textbf{Perception Module Ablation:} Comparison of baseline and our approach across different conditions. We report RMSE relative to \textit{same}. \textbf{P-BB:} Post-Backbone, \textbf{P-BEV:} Pre-BEV}
\label{tab:feature-distance}
\resizebox{\linewidth}{!}{%
\begin{tabular}{lcccccccc}
\toprule
& \multicolumn{4}{c}{\textbf{Baseline (CAA Policy~\cite{Chen2025caapolicy})}} & \multicolumn{4}{c}{\textbf{\methodname (Ours)}} \\
\cmidrule(lr){2-5}\cmidrule(lr){6-9}
& P-BB$\downarrow$ & P-BEV$\downarrow$ & Depth$\downarrow$ & BEV$\downarrow$ 
& P-BB$\downarrow$ & P-BEV$\downarrow$ & Depth$\downarrow$ & BEV$\downarrow$ \\
\midrule

Mild  & 1.67 & 0.40 & 0.69 & 0.17 
      & \textbf{0.54} & \textbf{0.27} & \textbf{0.69} & \textbf{0.07} \\
Large & 2.23 & 0.50 & 1.33 & 0.18  
      & \textbf{0.62} & \textbf{0.31} & \textbf{1.21} & \textbf{0.08} \\
Average & 1.95 & 0.45 & 1.01 & 0.18  
      & \textbf{0.58} & \textbf{0.29} & \textbf{0.95} & \textbf{0.08} \\
\bottomrule
\end{tabular}
} 
\vspace{-2em}
\end{table}

\section{Experimental Results and Discussion}
\label{sec:result}

In this section, we perform extensive experiments to validate our design's ability to enhance domain generalization for autonomous parking. We first introduce the experiment setup in \S\ref{sec:experiment_setup}, then discuss the experiment results of our solutions to the cross-domain gap: 1) VFMs for generalized surrounding perception (\S\ref{sec:res_perception}), 2) target relabeling for robust target tracking (\S\ref{sec:data_augmentation}), and 3) diffusion and Stanley control for motion (\S\ref{sec:res_diffusion_planning}). 4) performance trade-off in planner condition options (\S\ref{sec:representation_ablation}). Finally, we will show the potential of our method by a \textit{zero-shot} sim-to-real demonstration in a Gaussian-Splatting world (see \S\ref{sec:gaussian_splatting}).

\begin{table*}[t]
\centering
\caption{\textbf{Comprehensive Parking Performance Comparison:} Results comparing our method with the baseline and ablations of key pipeline components. Our modular design—featuring a generalizable perception module with visual foundation models (VFM) and robust planning with diffusion models—consistently and significantly outperforms the baseline in cross-domain benchmarks.}
\label{tab:parking_comparison}
\resizebox{\textwidth}{!}{%
\begin{tabular}{lccccccccccc}
\toprule
 & \textbf{TSR}$\uparrow$ & \textbf{TFR}$\downarrow$ & \textbf{NTSR}$\downarrow$ & \textbf{NTFR}$\downarrow$ & \textbf{CR}$\downarrow$ & \textbf{TR}$\downarrow$ & \textbf{APE} $(m)$ $\downarrow$ & \textbf{AOE} (degree) $\downarrow$ & \textbf{APT} (s) $\downarrow$ & \textbf{AIT} (s) $\downarrow$ \\
\toprule
\multicolumn{5}{l}{\textbf{Baseline (CAA Policy~\cite{Chen2025caapolicy})}} \\
\cmidrule(lr){1-3}
Same  & 0.86 $\pm$ 0.19 & 0.04 $\pm$ 0.10 & 0.06 $\pm$ 0.13 & 0.00 $\pm$ 0.00 & 0.02 $\pm$ 0.06 & 0.01 $\pm$ 0.04 & 0.32 $\pm$ 0.07 & 0.48 $\pm$ 0.66 & 20.56 $\pm$ 1.80 & 0.09 $\pm$ 0.00 \\
Mild  & 0.39 $\pm$ 0.22 & 0.17 $\pm$ 0.16 & \textbf{0.00 $\pm$ 0.00} & \textbf{0.00 $\pm$ 0.00} & 0.27 $\pm$ 0.20 & 0.18 $\pm$ 0.23 & 0.51 $\pm$ 0.21 & 1.53 $\pm$ 2.76 & 22.00 $\pm$ 3.13 & 0.09 $\pm$ 0.00 \\
Large & 0.32 $\pm$ 0.29 & 0.05 $\pm$ 0.08 & \textbf{0.00 $\pm$ 0.00} & 0.01 $\pm$ 0.04 & 0.39 $\pm$ 0.20 & 0.23 $\pm$ 0.25 & 0.39 $\pm$ 0.11 & 0.31 $\pm$ 0.66 & 21.90 $\pm$ 1.84 & 0.09 $\pm$ 0.00 \\
\toprule

\multicolumn{5}{l}{\textbf{Unfrozen Dino to Control Signal (CAA with unfrozen Dino)}} \\
\cmidrule(lr){1-5}
Same  & 0.78 $\pm$ 0.23 & \textbf{0.00 $\pm$ 0.00} & 0.06 $\pm$ 0.10 & 0.02 $\pm$ 0.06 & 0.06 $\pm$ 0.10 & 0.05 $\pm$ 0.10 & 0.36 $\pm$ 0.07 & 0.73 $\pm$ 0.83 & 22.41 $\pm$ 1.85 & 0.09 $\pm$ 0.00 \\
Mild  & 0.51 $\pm$ 0.31 & 0.13 $\pm$ 0.17 & 0.03 $\pm$ 0.07 & 0.02 $\pm$ 0.06 & 0.14 $\pm$ 0.20 & 0.17 $\pm$ 0.16 & 0.56 $\pm$ 0.13 & 1.01 $\pm$ 1.34 & 20.77 $\pm$ 2.79 & 0.09 $\pm$ 0.00 \\
Large & 0.89 $\pm$ 0.13 & 0.03 $\pm$ 0.07 & 0.02 $\pm$ 0.06 & 0.01 $\pm$ 0.04 & 0.02 $\pm$ 0.06 & 0.03 $\pm$ 0.07 & 0.33 $\pm$ 0.08 & 0.77 $\pm$ 0.95 & 21.33 $\pm$ 1.21 & 0.09 $\pm$ 0.00 \\
\toprule

\multicolumn{5}{l}{\textbf{Frozen Dino to Control Signal (CAA with frozen Dino)}} \\
\cmidrule(lr){1-5}
Same  & 0.84 $\pm$ 0.18 & 0.01 $\pm$ 0.04 & \textbf{0.00 $\pm$ 0.00} & \textbf{0.00 $\pm$ 0.00} & 0.07 $\pm$ 0.12 & 0.07 $\pm$ 0.09 & 0.39 $\pm$ 0.10 & 0.98 $\pm$ 1.05 & \textbf{20.27 $\pm$ 1.99} & \textbf{0.08 $\pm$ 0.00} \\
Mild  & 0.86 $\pm$ 0.15 & 0.06 $\pm$ 0.10 & \textbf{0.00 $\pm$ 0.00} & \textbf{0.00 $\pm$ 0.00} & 0.04 $\pm$ 0.07 & 0.03 $\pm$ 0.07 & 0.37 $\pm$ 0.07 & 0.50 $\pm$ 0.48 & \textbf{19.03 $\pm$ 1.74} & \textbf{0.08 $\pm$ 0.00} \\
Large & 0.78 $\pm$ 0.23 & 0.04 $\pm$ 0.07 & \textbf{0.00 $\pm$ 0.00} & \textbf{0.00 $\pm$ 0.00} & 0.08 $\pm$ 0.15 & 0.09 $\pm$ 0.12 & 0.37 $\pm$ 0.08 & 0.65 $\pm$ 0.76 & \textbf{19.83 $\pm$ 1.14} & \textbf{0.08 $\pm$ 0.00} \\
\toprule

\multicolumn{5}{l}{\textbf{\methodname-emb+seg (Ours)}} \\
\cmidrule(lr){1-3}
Same  & 0.92 $\pm$ 0.11 & 0.04 $\pm$ 0.07 & \textbf{0.00 $\pm$ 0.00} & \textbf{0.00 $\pm$ 0.00} & \textbf{0.00 $\pm$ 0.00} & 0.04 $\pm$ 0.10 & 0.30 $\pm$ 0.03 & \textbf{0.22 $\pm$ 0.04} & 21.60 $\pm$ 0.45 & 0.12 $\pm$ 0.00 \\
Mild  & 0.91 $\pm$ 0.09 & 0.07 $\pm$ 0.09 & \textbf{0.00 $\pm$ 0.00} & \textbf{0.00 $\pm$ 0.00} & \textbf{0.00 $\pm$ 0.00} & 0.02 $\pm$ 0.06 & \textbf{0.30 $\pm$ 0.03} & \textbf{0.22 $\pm$ 0.06} & 21.63 $\pm$ 0.48 & 0.12 $\pm$ 0.00 \\
Large & 0.94 $\pm$ 0.10 & 0.03 $\pm$ 0.07 & \textbf{0.00 $\pm$ 0.00} & \textbf{0.00 $\pm$ 0.00} & \textbf{0.00 $\pm$ 0.00} & 0.03 $\pm$ 0.07 & 0.31 $\pm$ 0.03 & \textbf{0.22 $\pm$ 0.05} & 21.90 $\pm$ 0.39 & 0.12 $\pm$ 0.00 \\

\toprule
\multicolumn{5}{l}{\textbf{\methodname-seg (Ours)}} \\
\cmidrule(lr){1-3}
Same  & \textbf{1.00 $\pm$ 0.00} & \textbf{0.00 $\pm$ 0.00} & \textbf{0.00 $\pm$ 0.00} & \textbf{0.00 $\pm$ 0.00} & \textbf{0.00 $\pm$ 0.00} & \textbf{0.00 $\pm$ 0.00} & \textbf{0.28 $\pm$ 0.06} & 0.87 $\pm$ 0.60 & 21.24 $\pm$ 0.61 & 0.11 $\pm$ 0.00 \\
Mild  & \textbf{1.00 $\pm$ 0.00} & \textbf{0.00 $\pm$ 0.00} & \textbf{0.00 $\pm$ 0.00} & \textbf{0.00 $\pm$ 0.00} & \textbf{0.00 $\pm$ 0.00} & \textbf{0.00 $\pm$ 0.00} & 0.30 $\pm$ 0.06 & 0.90 $\pm$ 0.50 & 21.40 $\pm$ 0.73 & 0.12 $\pm$ 0.00 \\
Large & \textbf{1.00 $\pm$ 0.00} & \textbf{0.00 $\pm$ 0.00} & \textbf{0.00 $\pm$ 0.00} & \textbf{0.00 $\pm$ 0.00} & \textbf{0.00 $\pm$ 0.00} & \textbf{0.00 $\pm$ 0.00} & \textbf{0.29 $\pm$ 0.06} & 1.02 $\pm$ 0.62 & 21.47 $\pm$ 0.66 & 0.12 $\pm$ 0.00 \\
\bottomrule
\end{tabular}
} 
\vspace{-2em}
\end{table*}

\subsection{Experiment Setup}
\label{sec:experiment_setup}

Our method is primarily compared with the state-of-the-art E2E approach, CAA-Policy~\cite{Chen2025caapolicy}, which demonstrates significant improvements over prior work~\cite{E2EAPA} and has released its training data. All evaluation pipelines are trained on the same dataset and tested under identical configurations.

For training, we use the dataset of 800 expert demonstration trajectories collected in CARLA \cite{Dosovitskiy17carla} following the description of \cite{E2EAPA} but with automated method given ground-truth BEV segmentation. 
Each trajectory contains camera images, depth images, segmentation maps, sensor information, vehicle pose, and controls. Our pipeline is fully trained on the expert dataset offline without any additional data collection and test in the three CARLA weather types: \textit{same}, \textit{mild} and \textit{large}, which means the same with the training setting, mild domain gap, and large domain gap respectively. We control the cloudiness, sun azimuth angle, sun altitude angle, fog density, wetness, precipitation and precipitation deposits of each configuration as follows:
\begin{itemize}
    \item Same: (15.0, 0.0, 75.0, 0.0, 0.0, 0.0, 0.0)
    \item Mild: (35.0, 20.0, 25.0, 2.0, 20.0, 30.0, 20.0)
    \item Large: (45.0, 30.0, 5.0, 3.0, 30.0, 45.0, 30.0)
\end{itemize}

At each configuration, we use 16 slots on each side of the parking to test our pipelines in CARLA Town04 map and start from 6 different initial pose for each slot. We follow the metrics in~\cite{E2EAPA}, including target success rate (TSR), target failure (TFR), non-target success rate (NTSR), non-target failure rate (NTFR), collision rate (CR), timeout rate (TR), average position error (APE), average orientation error (AOE), average parking time (APT) and average inference time (AIT).

\subsection{VFM Helps Visual Generalization}
\label{sec:res_perception}
We conduct a set of ablations to qualitatively and quantitatively show how the VFM leads to better cross-domain performance than previous E2E training \cite{Chen2025caapolicy}. We use DINOv2 ViT-B/14 distilled (86M parameters) as the backbone to process input camera images. We fix the vehicle on a point along the parking path, change the CARLA settings, and collect the example camera images in \textit{same}, \textit{mild} and \textit{large} settings (shown in Figure \ref{fig:robust_perception}). We extract the post-backbone image features and pre-BEV image features (post-backbone features will pass through some trainable CNN layers before the BEV view transform) and visualize these latent embeddings from E2E baseline \cite{Chen2025caapolicy} and our approach using UMap~\cite{2018arXivUMAP} with the same kernels. We find that 1) the DINO post-backbone feature is very consistent across domains, and can be neatly abstracted by the following CNN to provide a robust embedding to generate the UMap. In comparison, the EfficientNet \cite{tan2019efficientnet} used in E2E training post-backbone feature is much more sensitive to the visual disturbance and becomes more noisy in pre-BEV features. The same result can be read from Table \ref{tab:feature-distance} where we show the root-mean-square error (RMSE) in each map (Post-Backbone, Pre-BEV, depth and BEV) between the changed environment and the original. We find our VFM method can achieve 70.26\%, 35.56\%, 5.94\% and 55.56\% smaller domain gap respectively.

Then we conduct the close-loop parking research with frozen and unfrozen DINOv2 model and remain all other modules same with E2E parking \cite{Chen2025caapolicy} (see Table \ref{tab:parking_comparison} first three blocks). When we train the whole pipeline E2E, the perception is overfit to the training dataset or becomes unstable across all tasks. This could be supported by success rate where the Freezing DINO to Control outperforms the E2E and unfrozen option with 57.96\% and 13.76\% respectively.

\subsection{Data Augmentation for Robust Target Tracking}
\label{sec:data_augmentation}

\begin{figure}
    \centering
    \includegraphics[width=0.95\linewidth]{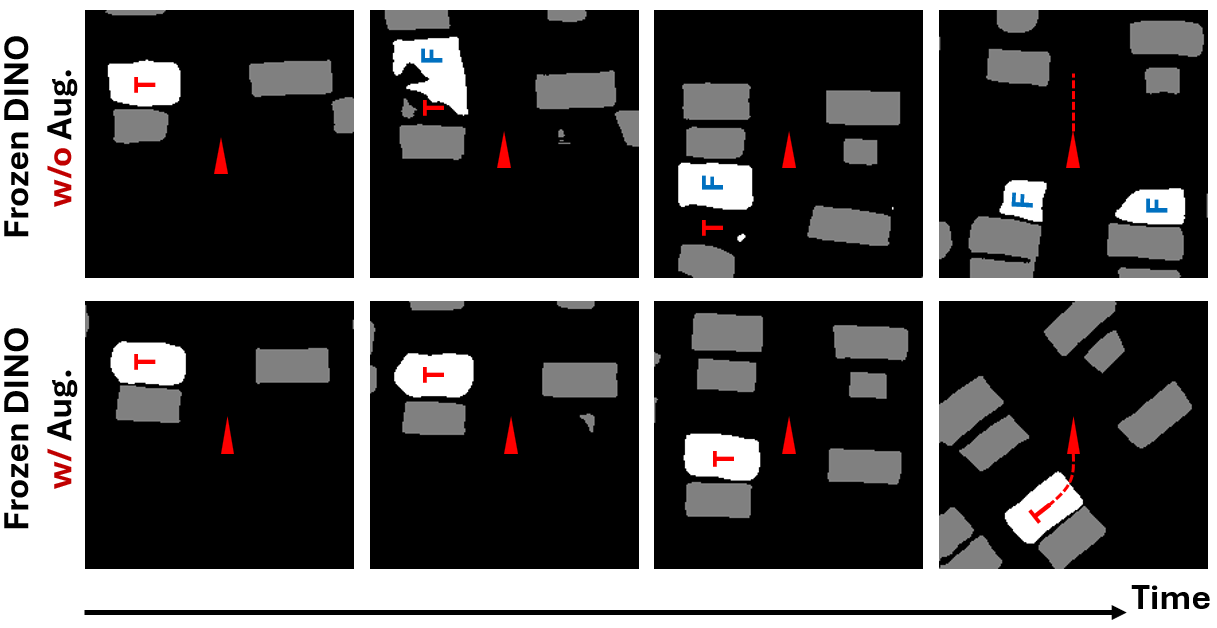}
    \caption{\textbf{Segmentation Map Along the Parking Trajectory:} Here we compare the segmentation map to represent the ego-centric perception where the red triangle is the vehicle. White and gray region means target and occupied slots, and true target and false target are indicated by T and F.}
    \label{fig:res_seg_cmp}
    \vspace{-5mm}
\end{figure}

While the generalized BEV has been derived in \S\ref{sec:res_perception}, it is also equally important to accurately merge the target information into the generated BEV. Under domain shifts, however, the target slot often drifts over time and causes planning failure. We address this with a simple yet effective data augmentation strategy that stabilizes target recognition. Using the same frozen DINO backbone, we compare performance with and without augmentation on a \textit{mild} cross-domain task. As shown in Figure~\ref{fig:res_seg_cmp}, the baseline drifts along the vehicle trajectory and fails to park, while our method maintains a consistent target slot and succeeds. This demonstrates that augmentation decouples target recognition from domain specific visual clues and expert dataset motion patterns, improving robustness across domains.

\begin{figure*}
    \centering
    \includegraphics[width=0.8\linewidth]{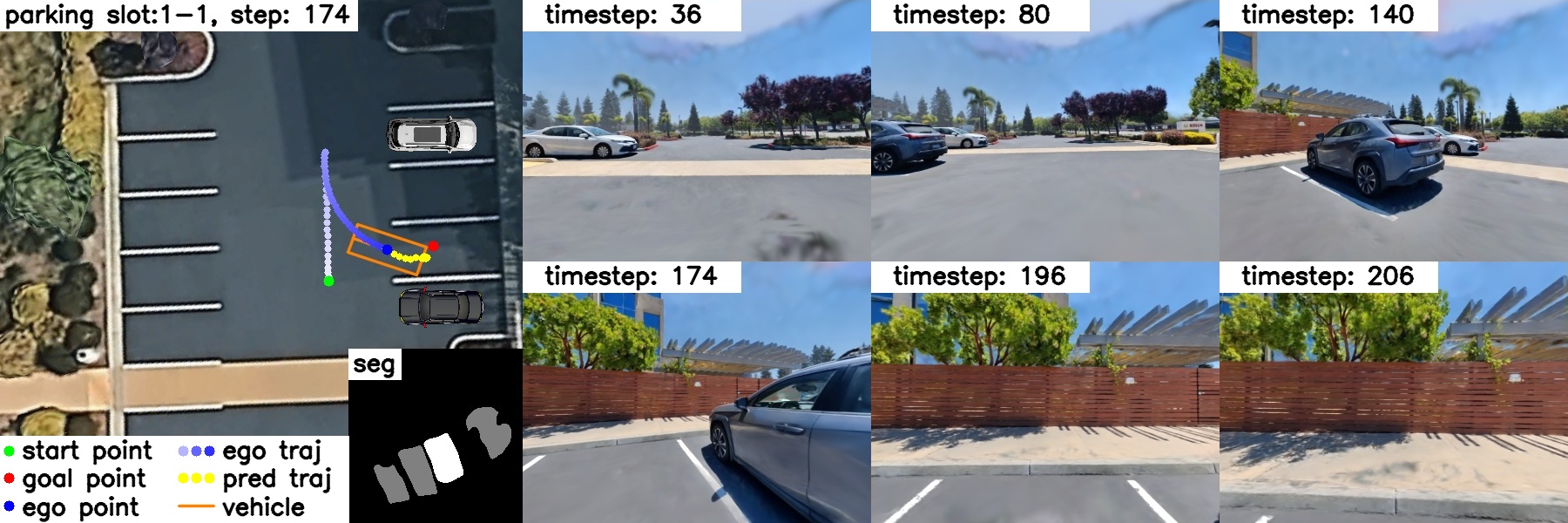}
    \caption{\textbf{Zero-shot GS Demonstration:} Our pipeline, trained exclusively in CARLA, is deployed in a 3DGS environment reconstructed from a real-world parking lot. The left panel shows the BEV of the parking lot, together with its segmentation, the ego trajectory (light blue), and the planned trajectory (yellow). Note that the BEV image is from Google Maps with vehicles models for illustration. The right panel presents back-camera views at different timesteps of a parking sequence.}
    \label{fig:gs_demo}
    \vspace{-5mm}
\end{figure*}

\subsection{Benefits of Diffusion Motion Planning with Pose Control}
\label{sec:res_diffusion_planning}
With the domain gap significantly reduced in perception, we aim to further decrease the compounding error by a combination of diffusion planning and Stanley pose control instead of one-step regression for the motion.  We replace the transformer motion head in \cite{Chen2025caapolicy} with the diffusion and Stanley control, then show the benchmark results at DDP-emb+seg block in Table \ref{tab:parking_comparison}. We find that our planning and control combination with the DINO embeddings can obviously outperform the CAA with DINO in the success rate, especially in the large domain gap scenario where it improves 16\%. Importantly, the APE and AOE of our method is also better, which shows our motion planning and control can lead to a higher precision in parking.

\subsection{Intermediate Representation Options Ablation}
\label{sec:representation_ablation}
We then study the simplification of the embeddings that the diffusion planning is conditioning: we replace the concatenation of the BEV embeddings and the segmentation into the segmentation only condition (last block in Table \ref{tab:parking_comparison}). As a result, we find it is a trade-off between the success rate and the parking precision: the segmentation only method reaches the ceiling TSR with 100\% across all tasks but subject to an unstable APE and worse AOE. The result is intuitive since it is easier to learn a useful pattern with a simplified condition, but its error also easily propagates into the planning.

\subsection{Real-world Gaussian Splatting Evaluation}
\label{sec:gaussian_splatting}

To further assess the limit of our proposed parking pipeline, we perform an extreme evaluation by testing it \textit{zero-shot} in a 3DGS environment~\cite{kerbl3Dgaussians} reconstructed from a real-world parking lot. Specifically, we collected thousands of images of a real parking scene and trained a hierarchical 3DGS model~\cite{hierarchicalgaussians24}. The resulting reconstruction, generated from dense image trajectories, enables free navigation within the parking area with minimal visual artifacts on relevant objects (aside from under-covered regions such as the sky). We use the same sensor setup in the 3DGS environment as in the CARLA simulator to ensure consistency. 

Despite the significant domain shift from CARLA to 3DGS, we observe non-trivial transferability. As shown in Figure~\ref{fig:gs_demo}, the BEV segmentation is able to capture the coarse structure of obstacles (gray) and target spot (white). Based on this perception, the planner produces a trajectory (yellow) that connects the agent’s current location (blue) to the designated parking spot (red), sometimes successfully parking the vehicle. These successful cases demonstrate that our pipeline can transfer—at least partially—across a large sim-to-real gap. However, failures remain common, often due to noisy segmentation or out-of-distribution layouts. Still, this experiment highlights an encouraging step toward bridging the substantial gap between simulated and real-world parking environments.

\section{Conclusion and future work}
In this paper, we design a novel autonomous parking pipeline integrating the visual foundation model for generalized perception, data relabeling for robust target fusion, and a diffusion model for consistent motion planning. Comprehensive benchmarks in cross-domain CARLA settings and 3DGS world environment highlight the advantages of our modular designs in bridging domain gaps where prior E2E approaches remain limited, marking a promising step toward generalizable autonomous parking. 

Ultimately, this line of research aims to close sim-to-real domain gap in autonomous driving/parking tasks without increasing data collection cost.
Future work may explore: 1) leveraging video world models to further reduce the domain gap between the CARLA and the real-world. 
2) collecting human demonstrations in 3DGS environment to train the model, and 
3) deploying the pipeline on real vehicles to validate performance across diverse scenarios.
\balance
\bibliographystyle{IEEEtran}
\bibliography{References}

\begin{thebibliography}{10}
\providecommand{\url}[1]{#1}
\csname url@rmstyle\endcsname
\providecommand{\newblock}{\relax}
\providecommand{\bibinfo}[2]{#2}
\providecommand\BIBentrySTDinterwordspacing{\spaceskip=0pt\relax}
\providecommand\BIBentryALTinterwordstretchfactor{4}
\providecommand\BIBentryALTinterwordspacing{\spaceskip=\fontdimen2\font plus
\BIBentryALTinterwordstretchfactor\fontdimen3\font minus \fontdimen4\font\relax}
\providecommand\BIBforeignlanguage[2]{{%
\expandafter\ifx\csname l@#1\endcsname\relax
\typeout{** WARNING: IEEEtran.bst: No hyphenation pattern has been}%
\typeout{** loaded for the language `#1'. Using the pattern for}%
\typeout{** the default language instead.}%
\else
\language=\csname l@#1\endcsname
\fi
#2}}

\bibitem{statista2025autonomous}
\BIBentryALTinterwordspacing
Statista, ``Autonomous vehicles worldwide - statistics \& facts,'' 2025. [Online]. Available: \url{https://www.statista.com/topics/3573/autonomous-vehicle-technology/}
\BIBentrySTDinterwordspacing

\bibitem{nhtsa2025automated}
\BIBentryALTinterwordspacing
NHTSA, ``Automated vehicles for safety,'' National Highway Traffic Safety Administration, 2025. [Online]. Available: \url{https://www.nhtsa.gov/vehicle-safety/automated-vehicles-safety}
\BIBentrySTDinterwordspacing

\bibitem{zhang2024reward}
H.~Zhang \emph{et~al.}, ``Reward-augmented reinforcement learning for continuous control in precision autonomous parking via policy optimization methods,'' \emph{arXiv preprint arXiv:2507.19642}, 2024.

\bibitem{aaa2016parking}
\BIBentryALTinterwordspacing
A.~R. Center, ``Avoid a vehicle collision when facing holiday shopping crowds,'' American Automobile Association, 2017. [Online]. Available: \url{https://www.ace.aaa.com/automotive/advocacy/parking-lot-collisions.html}
\BIBentrySTDinterwordspacing

\bibitem{zhao2024automatic}
Y.~Zhao, ``Automatic parking planning control method based on improved a* algorithm,'' \emph{arXiv preprint arXiv:2406.15429}, 2024.

\bibitem{leu2022improved}
J.~Leu, Y.~Wang, M.~Tomizuka, and S.~Di~Cairano, ``Improved a-search guided tree for autonomous trailer planning,'' in \emph{2022 IEEE/RSJ International Conference on Intelligent Robots and Systems (IROS)}.\hskip 1em plus 0.5em minus 0.4em\relax IEEE, 2022, pp. 7190--7196.

\bibitem{leu2022autonomous}
------, ``Autonomous vehicle parking in dynamic environments: An integrated system with prediction and motion planning,'' in \emph{2022 International Conference on Robotics and Automation (ICRA)}.\hskip 1em plus 0.5em minus 0.4em\relax IEEE, 2022, pp. 10\,890--10\,897.

\bibitem{shen2021collision}
X.~Shen, E.~L. Zhu, Y.~R. St{\"u}rz, and F.~Borrelli, ``Collision avoidance in tightly-constrained environments without coordination: a hierarchical control approach,'' in \emph{2021 IEEE International Conference on Robotics and Automation (ICRA)}.\hskip 1em plus 0.5em minus 0.4em\relax IEEE, 2021, pp. 2674--2680.

\bibitem{yang2021towards}
Y.~Yang, M.~Pan, S.~Jiang, J.~Wang, W.~Wang, J.~Wang, and M.~Wang, ``Towards autonomous parking using vision-only sensors,'' in \emph{2021 IEEE/RSJ International Conference on Intelligent Robots and Systems (IROS)}.\hskip 1em plus 0.5em minus 0.4em\relax IEEE, 2021, pp. 2038--2044.

\bibitem{philion2020lift}
J.~Philion and S.~Fidler, ``Lift, splat, shoot: Encoding images from arbitrary camera rigs by implicitly unprojecting to 3d,'' in \emph{European conference on computer vision}.\hskip 1em plus 0.5em minus 0.4em\relax Springer, 2020, pp. 194--210.

\bibitem{li2024bevformer}
Z.~Li, W.~Wang, H.~Li, E.~Xie, C.~Sima, T.~Lu, Q.~Yu, and J.~Dai, ``Bevformer: learning bird's-eye-view representation from lidar-camera via spatiotemporal transformers,'' \emph{IEEE Transactions on Pattern Analysis and Machine Intelligence}, 2024.

\bibitem{rathour2018vision}
S.~Rathour, V.~John, M.~Nithilan, and S.~Mita, ``Vision and dead reckoning-based end-to-end parking for autonomous vehicles,'' in \emph{2018 IEEE Intelligent Vehicles Symposium (IV)}.\hskip 1em plus 0.5em minus 0.4em\relax IEEE, 2018, pp. 2182--2187.

\bibitem{hu2022st}
S.~Hu, L.~Chen, P.~Wu, H.~Li, J.~Yan, and D.~Tao, ``St-p3: End-to-end vision-based autonomous driving via spatial-temporal feature learning,'' in \emph{European Conference on Computer Vision}.\hskip 1em plus 0.5em minus 0.4em\relax Springer, 2022, pp. 533--549.

\bibitem{E2EAPA}
Y.~Yang, D.~Chen, T.~Qin, X.~Mu, C.~Xu, and M.~Yang, ``E2e parking: Autonomous parking by the end-to-end neural network on the carla simulator,'' in \emph{Conference on IEEE Intelligent Vehicles Symposium}, 2024.

\bibitem{Chen2025caapolicy}
C.~Chen, S.~Yao, Y.~He, T.~Feng, R.~Song, Y.~Guo, X.~Huang, C.~Wu, R.~Liu, and C.~Feng, ``End-to-end visual autonomous parking via control-aided attention,'' \emph{arXiv preprint arXiv:2509.11090}, 2025.

\bibitem{mueller2018driving}
M.~Mueller, A.~Dosovitskiy, B.~Ghanem, and V.~Koltun, ``Driving policy transfer via modularity and abstraction,'' in \emph{Conference on Robot Learning}.\hskip 1em plus 0.5em minus 0.4em\relax PMLR, 2018, pp. 1--15.

\bibitem{codevilla2019exploring}
F.~Codevilla, E.~Santana, A.~Lopez, and A.~Gaidon, ``Exploring the limitations of behavior cloning for autonomous driving,'' in \emph{2019 IEEE/CVF International Conference on Computer Vision (ICCV)}.\hskip 1em plus 0.5em minus 0.4em\relax IEEE, 2019, pp. 9328--9337.

\bibitem{domain_randomization}
G.~Zhan, Y.~Lyu, S.~E. Li, Y.~Jiang, X.~Zhang, and L.~Tao, ``Enhance generality by model-based reinforcement learning and domain randomization,'' in \emph{2023 7th CAA International Conference on Vehicular Control and Intelligence (CVCI)}, 2023, pp. 1--6.

\bibitem{yasarla2025roca}
R.~Yasarla, S.~Han, H.-P. Cheng, L.~Liu, S.~Mahajan, A.~Bhattacharyya, Y.~Shi, R.~Garrepalli, H.~Cai, and F.~Porikli, ``Roca: Robust cross-domain end-to-end autonomous driving,'' \emph{arXiv preprint arXiv:2506.10145}, 2025.

\bibitem{dosovitskiy2020image}
A.~Dosovitskiy, L.~Beyer, A.~Kolesnikov, D.~Weissenborn, X.~Zhai, T.~Unterthiner, M.~Dehghani, M.~Minderer, G.~Heigold, S.~Gelly, \emph{et~al.}, ``An image is worth 16x16 words: Transformers for image recognition at scale,'' \emph{arXiv preprint arXiv:2010.11929}, 2020.

\bibitem{caron2021dino}
M.~Caron, H.~Touvron, I.~Misra, H.~J{\'e}gou, J.~Mairal, P.~Bojanowski, and A.~Joulin, ``Emerging properties in self-supervised vision transformers,'' in \emph{Proceedings of the IEEE/CVF International Conference on Computer Vision}, 2021, pp. 9650--9660.

\bibitem{oquab2023dinov2}
M.~Oquab, T.~Darcet, T.~Moutakanni, H.~Vo, M.~Szafraniec, V.~Khalidov, P.~Fernandez, D.~Haziza, F.~Massa, A.~El-Nouby, \emph{et~al.}, ``Dinov2: Learning robust visual features without supervision,'' \emph{arXiv preprint arXiv:2304.07193}, 2023.

\bibitem{kirillov2023segment}
A.~Kirillov, E.~Mintun, N.~Ravi, H.~Mao, C.~Rolland, L.~Gustafson, T.~Xiao, S.~Whitehead, A.~C. Berg, W.-Y. Lo, \emph{et~al.}, ``Segment anything,'' in \emph{Proceedings of the IEEE/CVF international conference on computer vision}, 2023, pp. 4015--4026.

\bibitem{depthanything}
L.~Yang, B.~Kang, Z.~Huang, X.~Xu, J.~Feng, and H.~Zhao, ``Depth anything: Unleashing the power of large-scale unlabeled data,'' in \emph{CVPR}, 2024.

\bibitem{li2024parkinge2e}
C.~Li, Z.~Ji, Z.~Chen, T.~Qin, and M.~Yang, ``Parkinge2e: Camera-based end-to-end parking network, from images to planning,'' in \emph{2024 IEEE/RSJ International Conference on Intelligent Robots and Systems (IROS)}.\hskip 1em plus 0.5em minus 0.4em\relax IEEE, 2024, pp. 13\,206--13\,212.

\bibitem{olmos2025overview}
J.~S. Olmos~Medina, J.~G. Maradey~L{\'a}zaro, A.~Rass{\~o}lkin, and H.~Gonz{\'a}lez~Acu{\~n}a, ``An overview of autonomous parking systems: Strategies, challenges, and future directions,'' \emph{Sensors}, vol.~25, no.~14, p. 4328, 2025.

\bibitem{liang2022bevfusion}
T.~Liang, H.~Xie, K.~Yu, Z.~Xia, Z.~Lin, Y.~Wang, T.~Tang, B.~Wang, and Z.~Tang, ``Bevfusion: A simple and robust lidar-camera fusion framework,'' \emph{Advances in Neural Information Processing Systems}, vol.~35, pp. 10\,421--10\,434, 2022.

\bibitem{bojarski2016end}
M.~Bojarski, D.~Del~Testa, D.~Dworakowski, B.~Firner, B.~Flepp, P.~Goyal, L.~D. Jackel, M.~Monfort, U.~Muller, J.~Zhang, \emph{et~al.}, ``End to end learning for self-driving cars,'' \emph{arXiv preprint arXiv:1604.07316}, 2016.

\bibitem{pomerleau1988alvinn}
D.~A. Pomerleau, ``Alvinn: An autonomous land vehicle in a neural network,'' \emph{Advances in neural information processing systems}, vol.~1, 1988.

\bibitem{liu2022bevfusion}
Z.~Liu, H.~Tang, A.~Amini, X.~Yang, H.~Mao, D.~Rus, and S.~Han, ``Bevfusion: Multi-task multi-sensor fusion with unified bird's-eye view representation,'' in \emph{IEEE International Conference on Robotics and Automation (ICRA)}, 2023.

\bibitem{lin2024rcbevdet}
Z.~Lin, Z.~Liu, Z.~Xia, X.~Wang, Y.~Wang, S.~Qi, Y.~Dong, N.~Dong, L.~Zhang, and C.~Zhu, ``Rcbevdet: Radar-camera fusion in bird's eye view for 3d object detection,'' in \emph{2024 IEEE/CVF Conference on Computer Vision and Pattern Recognition (CVPR)}, 2024, pp. 14\,928--14\,937.

\bibitem{li2023domain}
J.~Li, R.~Xu, J.~Ma, Q.~Zou, J.~Ma, and H.~Yu, ``Domain adaptive object detection for autonomous driving under foggy weather,'' in \emph{Proceedings of the IEEE/CVF winter conference on applications of computer vision}, 2023, pp. 612--622.

\bibitem{Gupta2024robustweather}
H.~Gupta, O.~Kotlyar, H.~Andreasson, and A.~J. Lilienthal, ``Robust object detection in challenging weather conditions,'' in \emph{2024 IEEE/CVF Winter Conference on Applications of Computer Vision (WACV)}, 2024, pp. 7508--7517.

\bibitem{ross2010efficient}
S.~Ross and D.~Bagnell, ``Efficient reductions for imitation learning,'' in \emph{Proceedings of the thirteenth international conference on artificial intelligence and statistics}.\hskip 1em plus 0.5em minus 0.4em\relax JMLR Workshop and Conference Proceedings, 2010, pp. 661--668.

\bibitem{ross2011reduction}
S.~Ross, G.~Gordon, and D.~Bagnell, ``A reduction of imitation learning and structured prediction to no-regret online learning,'' in \emph{Proceedings of the fourteenth international conference on artificial intelligence and statistics}.\hskip 1em plus 0.5em minus 0.4em\relax JMLR Workshop and Conference Proceedings, 2011, pp. 627--635.

\bibitem{ho2016generative}
J.~Ho and S.~Ermon, ``Generative adversarial imitation learning,'' \emph{Advances in neural information processing systems}, vol.~29, 2016.

\bibitem{fu2017learning}
J.~Fu, K.~Luo, and S.~Levine, ``Learning robust rewards with adversarial inverse reinforcement learning,'' \emph{arXiv preprint arXiv:1710.11248}, 2017.

\bibitem{li2023learning}
C.~Li, M.~Vlastelica, S.~Blaes, J.~Frey, F.~Grimminger, and G.~Martius, ``Learning agile skills via adversarial imitation of rough partial demonstrations,'' in \emph{Conference on Robot Learning}.\hskip 1em plus 0.5em minus 0.4em\relax PMLR, 2023, pp. 342--352.

\bibitem{chen2021decision}
L.~Chen, K.~Lu, A.~Rajeswaran, K.~Lee, A.~Grover, M.~Laskin, P.~Abbeel, A.~Srinivas, and I.~Mordatch, ``Decision transformer: Reinforcement learning via sequence modeling,'' \emph{Advances in neural information processing systems}, vol.~34, pp. 15\,084--15\,097, 2021.

\bibitem{zhao2023learning}
T.~Z. Zhao, V.~Kumar, S.~Levine, and C.~Finn, ``Learning fine-grained bimanual manipulation with low-cost hardware,'' \emph{arXiv preprint arXiv:2304.13705}, 2023.

\bibitem{janner2022planning}
M.~Janner, Y.~Du, J.~Tenenbaum, and S.~Levine, ``Planning with diffusion for flexible behavior synthesis,'' in \emph{International Conference on Machine Learning}, 2022.

\bibitem{chi2023diffusionpolicy}
C.~Chi, S.~Feng, Y.~Du, Z.~Xu, E.~Cousineau, B.~Burchfiel, and S.~Song, ``Diffusion policy: Visuomotor policy learning via action diffusion,'' in \emph{Proceedings of Robotics: Science and Systems (RSS)}, 2023.

\bibitem{jiang2023motiondiffuser}
C.~Jiang, A.~Cornman, C.~Park, B.~Sapp, Y.~Zhou, D.~Anguelov, \emph{et~al.}, ``Motiondiffuser: Controllable multi-agent motion prediction using diffusion,'' in \emph{Proceedings of the IEEE/CVF conference on computer vision and pattern recognition}, 2023, pp. 9644--9653.

\bibitem{zheng2025diffusion}
Y.~Zheng, R.~Liang, K.~Zheng, J.~Zheng, L.~Mao, J.~Li, W.~Gu, R.~Ai, S.~E. Li, X.~Zhan, \emph{et~al.}, ``Diffusion-based planning for autonomous driving with flexible guidance,'' \emph{arXiv preprint arXiv:2501.15564}, 2025.

\bibitem{wu2025learningwheelchairtennisnavigation}
Z.~Wu, Z.~Zaidi, A.~Patil, Q.~Xiao, and M.~Gombolay, ``Learning wheelchair tennis navigation from broadcast videos with domain knowledge transfer and diffusion motion planning,'' in \emph{2025 IEEE International Conference on Robotics and Automation (ICRA)}.\hskip 1em plus 0.5em minus 0.4em\relax IEEE, 2025, pp. 4837--4844.

\bibitem{perez2018film}
E.~Perez, F.~Strub, H.~De~Vries, V.~Dumoulin, and A.~Courville, ``Film: Visual reasoning with a general conditioning layer,'' in \emph{Proceedings of the AAAI conference on artificial intelligence}, vol.~32, no.~1, 2018.

\bibitem{2018arXivUMAP}
L.~{McInnes}, J.~{Healy}, and J.~{Melville}, ``{UMAP: Uniform Manifold Approximation and Projection for Dimension Reduction},'' \emph{ArXiv e-prints}, Feb. 2018.

\bibitem{thrun2006stanley}
S.~Thrun, M.~Montemerlo, H.~Dahlkamp, D.~Stavens, A.~Aron, J.~Diebel, P.~Fong, J.~Gale, M.~Halpenny, G.~Hoffmann, \emph{et~al.}, ``Stanley: The robot that won the darpa grand challenge,'' \emph{Journal of field Robotics}, vol.~23, no.~9, pp. 661--692, 2006.

\bibitem{Dosovitskiy17carla}
A.~Dosovitskiy, G.~Ros, F.~Codevilla, A.~Lopez, and V.~Koltun, ``{CARLA}: {An} open urban driving simulator,'' in \emph{Proceedings of the 1st Annual Conference on Robot Learning}, 2017, pp. 1--16.

\bibitem{tan2019efficientnet}
M.~Tan and Q.~Le, ``Efficientnet: Rethinking model scaling for convolutional neural networks,'' in \emph{International conference on machine learning}.\hskip 1em plus 0.5em minus 0.4em\relax PMLR, 2019, pp. 6105--6114.

\bibitem{kerbl3Dgaussians}
B.~Kerbl, G.~Kopanas, T.~Leimk{\"u}hler, and G.~Drettakis, ``3d gaussian splatting for real-time radiance field rendering,'' \emph{ACM Transactions on Graphics}, vol.~42, no.~4, July 2023.

\bibitem{hierarchicalgaussians24}
B.~Kerbl, A.~Meuleman, G.~Kopanas, M.~Wimmer, A.~Lanvin, and G.~Drettakis, ``A hierarchical 3d gaussian representation for real-time rendering of very large datasets,'' \emph{ACM Transactions on Graphics}, vol.~43, no.~4, July 2024.

\end{thebibliography}

\end{document}